\documentclass[letterpaper, 10 pt, conference]{ieeeconf}  % Comment this line out if you need a4paper

\IEEEoverridecommandlockouts                              % This command is only needed if 
\usepackage[utf8x]{inputenc}
\usepackage{hyperref}
\usepackage{graphicx}
\usepackage{wrapfig}
\usepackage{url}
\usepackage[export]{adjustbox}
\usepackage{physics} 
\usepackage{breqn}
\usepackage{hyperref}
\usepackage{amssymb}
\usepackage{amsmath}
\usepackage{siunitx}
\graphicspath{{images/}}

        \usepackage{float}
         \usepackage{epsfig} 
         \usepackage{graphicx}
         \usepackage{graphics}
         \usepackage{latexsym} 
         \usepackage{amsmath}
         \usepackage{psfrag}
         \usepackage{amsmath}
         \usepackage{amssymb}
         \usepackage{multicol}
         \usepackage{wrapfig}
         \usepackage{cite}
         \usepackage{url}
         \usepackage{bm} % bold math
        \usepackage{subcaption}
        
\usepackage{mathptmx}
\usepackage{anyfontsize}
\usepackage{t1enc}
        
\usepackage[usenames, dvipsnames]{color}        
\title{\LARGE \bf
Nonlinear Dynamics of Binocular Rivalry: A Comparative Study}

\author{Yashaswini Murthy$^{1}$% <-this % stops a space
\thanks{$^{1}$Department of Mechanical Engineering, Indian Institute of Technology Bombay, India.}
}

\begin{document}

\maketitle
\thispagestyle{empty}
\pagestyle{empty}

%%%%%%%%%%%%%%%%%%%%%%%%%%%%%%%%%%%%%%%%%%%%%%%%%%%%%%%%%%%%%%%%%%%%%%%%%%%%%%%%
\begin{abstract}
When our eyes are presented with the same image, the brain processes it to view it as a single coherent one. The lateral shift in the position of our eyes, causes the two images to possess certain differences, which our brain exploits for the purpose of depth perception and to gauge the size of objects at different distances, a process commonly known as stereopsis. However, when presented with two different visual stimuli, the visual awareness alternates. This phenomenon of binocular rivalry is a result of competition between the corresponding neuronal populations of the two eyes. The article presents a comparative study of various dynamical models proposed to capture this process. It goes on to study the effect of a certain parameter on the rate of perceptual alternations and proceeds to disprove the initial propositions laid down to characterise this phenomenon. It concludes with a discussion on the possible future work that can be conducted to obtain a better picture of the neuronal functioning behind this rivalry. 
\end{abstract}

%%%%%%%%%%%%%%%%%%%%%%%%%%%%%%%%%%%%%%%%%%%%%%%%%%%%%%%%%%%%%%%%%%%%%%%%%%%%%%%%
\section{INTRODUCTION}

Binocular rivalry is the striking phenomenon that ensues when the two eyes view markedly different stimuli. The observer perceives only one stimulus at a time, and perception alternates between the two stimuli at irregular intervals\cite{levelt}. The perceived durations of the images are stochastic and uncorrelated with previous perceived durations\cite{kalar}. One of the initial studies conducted on this phenomenon was by Levelt in 1965. His four propositions went onto become the cornerstone of many of the mathematical models developed later on the same. They can be broadly summarised as:
\begin{enumerate}
    \item Increasing stimulus strength for one eye will increase the
perceptual predominance of that eye’s stimulus.
    \item Increasing stimulus strength for one eye will not affect the
average perceptual dominance duration of that eye’s stimulus.
Instead, it will reduce the average perceptual dominance duration of the other eye’s stimulus.
    \item Increasing stimulus strength for one eye will increase the
perceptual alternation rate.
    \item Increasing stimulus strength in both eyes while keeping it
equal between eyes will increase the perceptual alternation rate.
\end{enumerate}
This switching of dominance between the two perceptions is primarily due to the inhibition the dominant side exerts over the other. This reciprocal inhibition, is constantly accompanied by a slow negative feedback exerted by the suppressed side. This negative feedback acts through spike frequency adaptation and synaptic depression. \\
Spike frequency adaptation is a result of a decrease in the activity of the dominant side, eventually leading to the suppressed neuronal population expressing itself. Whereas synaptic depression results when the strength of the connectivity between neurons has reduced, thereby leading to decreased inhibitory effects of the dominant population on the suppressed one. These principles have been incorporated into developing various mathematical models, such as  \cite{kalar}, \cite{wilson}, \cite{laing}, \cite{main}, \cite{blake}, \cite{gberg}, \cite{lago}, \cite{lehky}, \cite{mats} and \cite{stol}. This article proceeds to study the dynamics predicted by Wilson's model, Laing and Chow's model and it's adaptation only variant as well as Kalarickal and Marshall's model.

\section{WILSON'S MODEL}
One of the distinguishing features of Wilson's model is the existence of separate neuronal populations to exert the inhibitory effect. That is, the neuronal population responsible for perception is physically different from the neuronal population exerting the inhibition on the suppressed side. The rivalry dynamics are modelled as below:

\begin{equation*}
    \tau\dot{E_1} = -E_1 + \frac{100(V_1-gI_2)_{+}^{2}}{(10+H_1)^2+(V_1-gI_2)_{+}^{2}}
\end{equation*}
\begin{equation*}
    \tau_{H}\dot{H_1} = -H_1 + hE_1
\end{equation*}
\begin{equation*}
    \tau_{I}\dot{I_1} = -I_1 + E_1
\end{equation*}
\begin{equation*}
    \tau\dot{E_2} = -E_2 + \frac{100(V_2-gI_1)_{+}^{2}}{(10+H_2)^2+(V_2-gI_1)_{+}^{2}}
\end{equation*}
\begin{equation*}
    \tau_{H}\dot{H_2} = -H_2 + hE_2
\end{equation*}
\begin{equation*}
    \tau_{I}\dot{I_2} = -I_2 + E_2
\end{equation*}
$V_i$ are the input stimuli, $g$ represents the cross inhibition parameter - it scales the strength of the inhibition exerted by the inhibitory neuronal population, $E_i$ represents the population firing rate, an indication of the activity of the neuronal population (higher firing rate implies greater dominance in perception), $I_i$ is the inhibitory firing rate of the $i$th inhibition population on the other, $H_i$ is the adaptation variable, $\tau$, $\tau_H$ and $\tau_I$ correspond to the time constant of the firing rate of the excitatory populations, adaption process and firing rate of the inhibitory population respectively \cite{wilson}. The asymptotic firing rates is determined by Naka-Rushton function, where  $(V_1-gI_2)_{+} = (V_1-gI_2)$ if $(V_1-gI_2) > 0$ else $V_1-gI_2)_{+} = 0$.
\\The above equations are simulated on MATLAB for various stimulus strengths and these parametric values: $g$ = 0.45, $\tau$ = 20$ms$, $\tau_H$ = 900$ms$ (adaptation is a relatively slow process), $\tau_I$ = 11$ms$ and $h$ = 0.47. The simulation results are as follows: 

\begin{figure}[H]
\centering
\includegraphics[width=8cm, height=2cm]{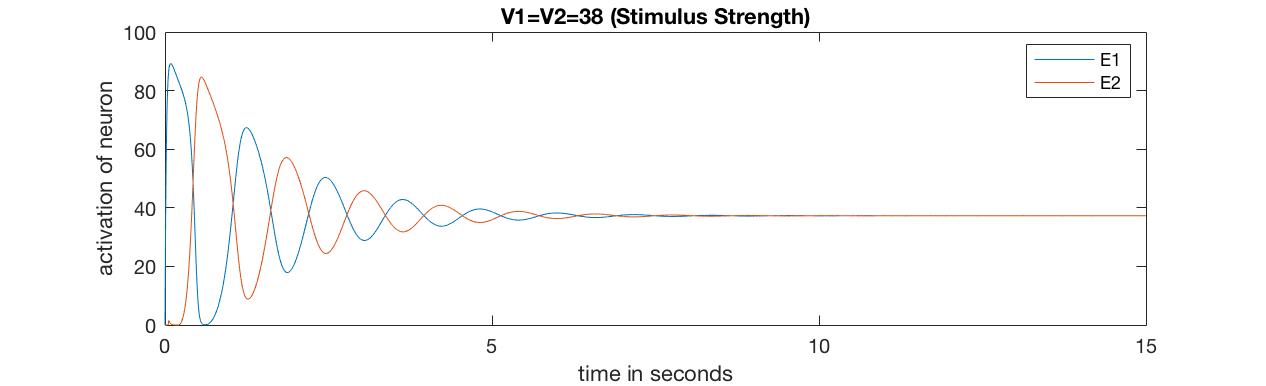}\vspace{2mm}
\includegraphics[width=8cm, height=2cm]{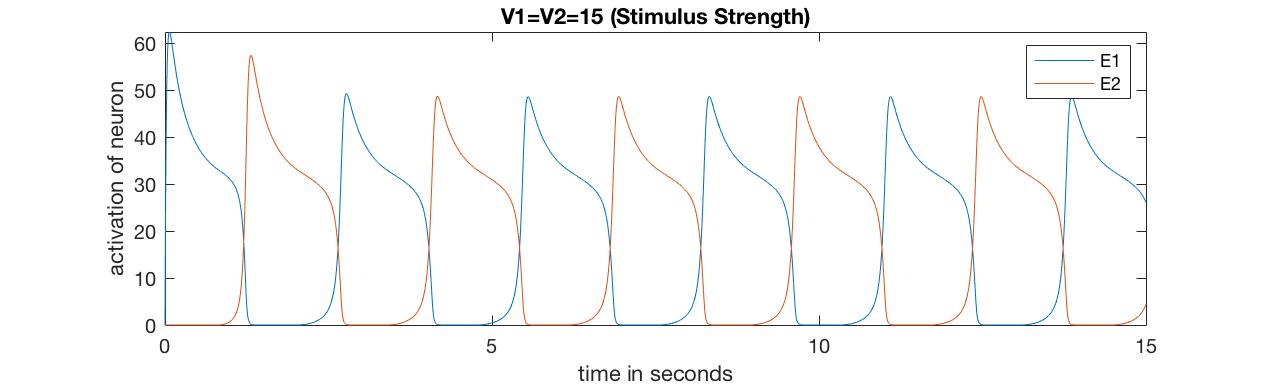}\vspace{2mm}
\includegraphics[width=8cm, height=2cm]{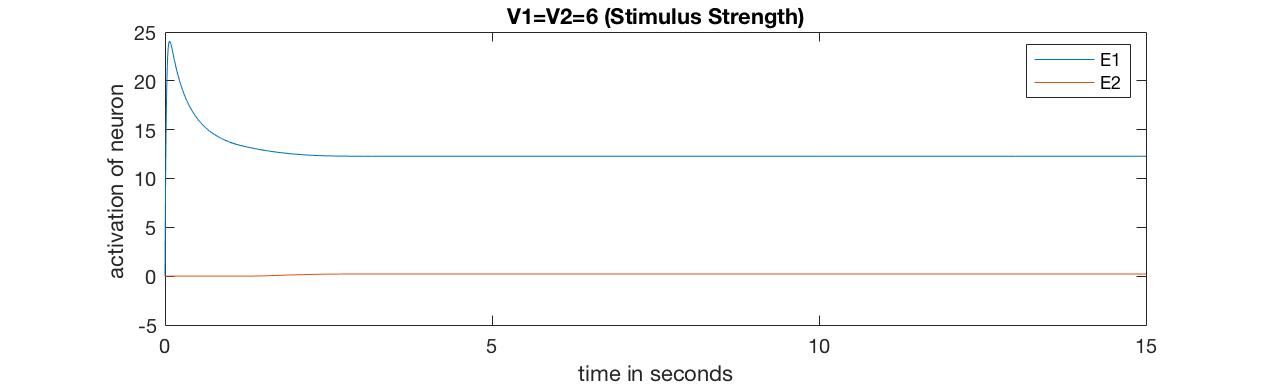}\vspace{2mm}
\includegraphics[width=8cm, height=2cm]{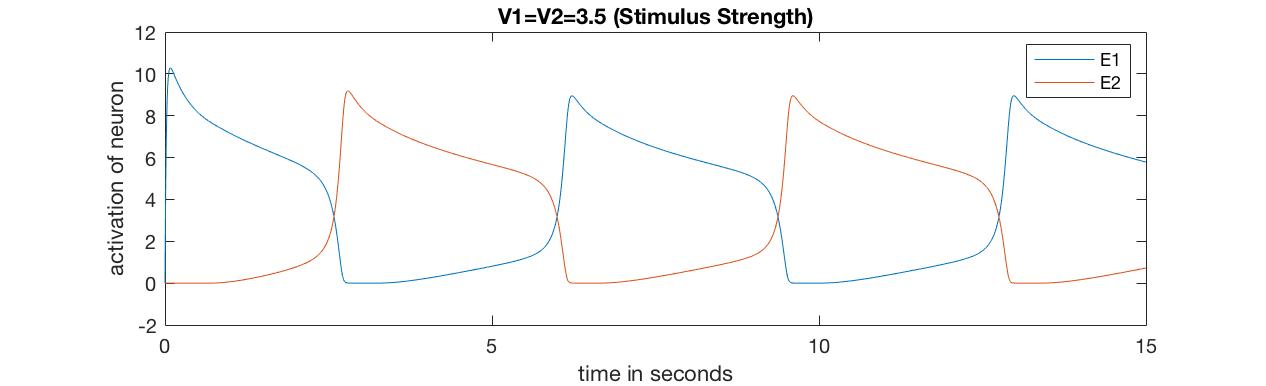}\vspace{2mm}
\includegraphics[width=8cm, height=2cm]{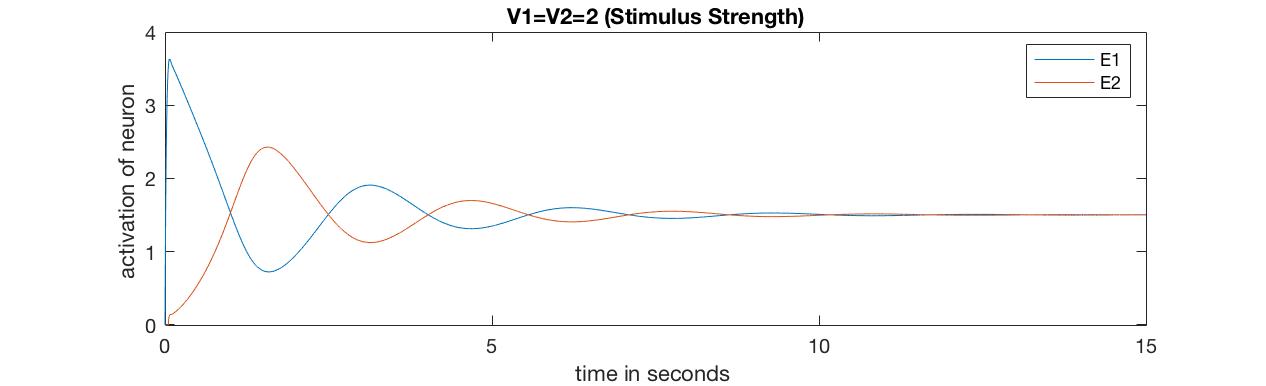}
\caption{Wilson Model: Variation of perceptual alternation with stimulus strength.}
\label{fig:pic7}
\end{figure}

At high stimulus strengths, followed by an initial transient it is seen that both the stimuli equally dominate. There is an absence of alternation. However as the stimuli strength decreases, there is an onset of oscillations. The mean dominance time of these oscillations vary with stimulus strength. Increasing the stimulus strength decreases the mean dominance time of these oscillations, which is in compliance with the fourth proposition of Levelt, which states that increasing the stimulus strength while keeping them equal, increases the perceptual alternation rate. Upon further reducing the stimulus strength, there is a clear dominance of stimuli over the other, solely based on the initial conditions. There is no oscillation. This is commonly referred to as the "winner take all" regime. At an even lower stimulus strength, there is again an onset of perceptual oscillations. However these oscillations are different as when compared to the second regime. As the stimulus strength increases there is an increase in the mean dominance time of these oscillations, completely in contrast to what was stated in the fourth proposition. Upon further reduction in stimuli strength, it is seen that there is a fusion of these stimuli perceptions again. Hence it is evident that there is a strong absence of monotonous relationship between stimulus strength and the perceptual alternations. This is in strong contradiction of the initial fourth proposition of Levelt.
Although these simulations are for a certain set of parametric values, the nonmonotonicity of this behaviour is robust across many such feasible parametric values \cite{main}. \\
%A similar such behaviour is depicted by Laing and Chow's model as well. 
\vspace{-2mm}
\section{LAING AND CHOW'S MODEL}
In contrast to the above Wilson's model, Laing and Chow's dynamics describes a single neuronal population to exert the inhibitory effect and result in perception. Unlike other models, Laing and Chow's model also describes the evolution of adaptation and synaptic depression as first order equations.
\begin{equation*}
    \dot{u_1} = -u_1 + f(\alpha u_1g_1 - \beta u_2g_2 - a_1 +I_1)
\end{equation*}
\begin{equation*}
    \tau_{a}\dot{a_1} = -a_1 + \phi_a f(\alpha u_1g_1 - \beta u_2g_2 - a_1 +I_1)
\end{equation*}
\begin{equation*}
    \tau_{d}\dot{g_1} = 1 - g_1 - g_1\phi_d f(\alpha u_1g_1 - \beta u_2g_2 - a_1 +I_1)
\end{equation*}
\begin{equation*}
    \dot{u_2} = -u_2 + f(\alpha u_2g_2 - \beta u_1g_1 - a_2 +I_2)
\end{equation*}
\begin{equation*}
    \tau_{a}\dot{a_2} = -a_2 + \phi_a f(\alpha u_2g_2 - \beta u_1g_1 - a_2 +I_2)
\end{equation*}
\begin{equation*}
    \tau_{d}\dot{g_2} = 1 - g_2 - g_2\phi_d f(\alpha u_2g_2 - \beta u_1g_1 - a_2 +I_2)
\end{equation*}
$I_i$ are the input stimuli, $\beta$ represents the cross inhibition parameter, $u_i$ represents the average population firing rate, an indication of the activity of the neuronal population, $g_i$ is the synaptic depression, $a_i$ is the adaptation variable, $\tau_a$ and $\tau_d$ correspond to the time constant of adaption and synaptic depression processes respectively \cite{laing}. The gain function $f$ is taken to be the Heaviside step function, i.e., $f(x) = 1$ for $x ≥ 0$ and $f(x) = 0$ for $x < 0$. $\alpha$, $\phi_a$ and $\phi_d$ are scaling parameters. All the dynamic variables are normalised to hold values between $0$ and $1$. \\
The following simulation results are obtained for these parametric values: $\alpha = 0.35$, $\phi_a = 0.6$, $\phi_d = 0.6$, $\beta = 0.7$, $\tau_a$ = 20$ms$ and $\tau_d$ = 40$ms$.

\begin{figure}[H]
\centering
\includegraphics[width=8cm, height=2cm]{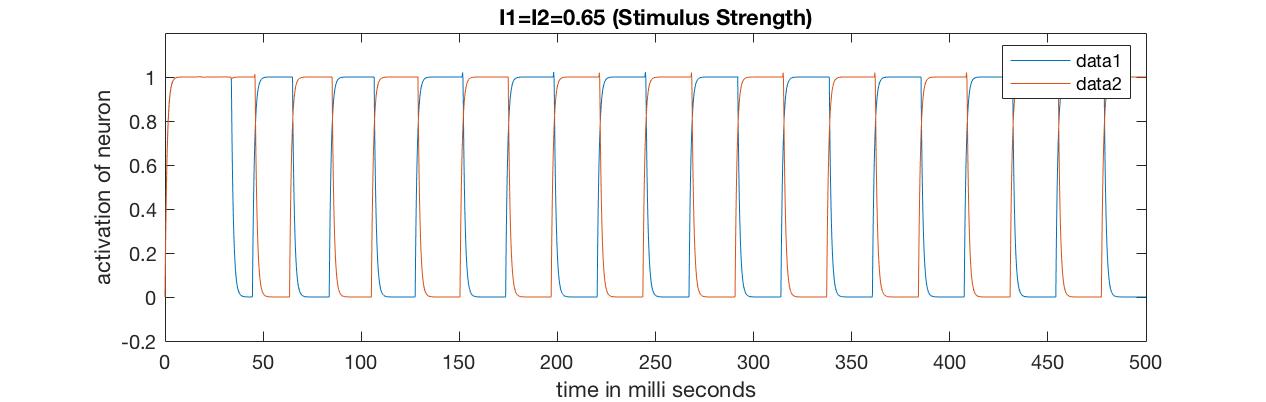}\vspace{1.5mm}
\includegraphics[width=8cm, height=2cm]{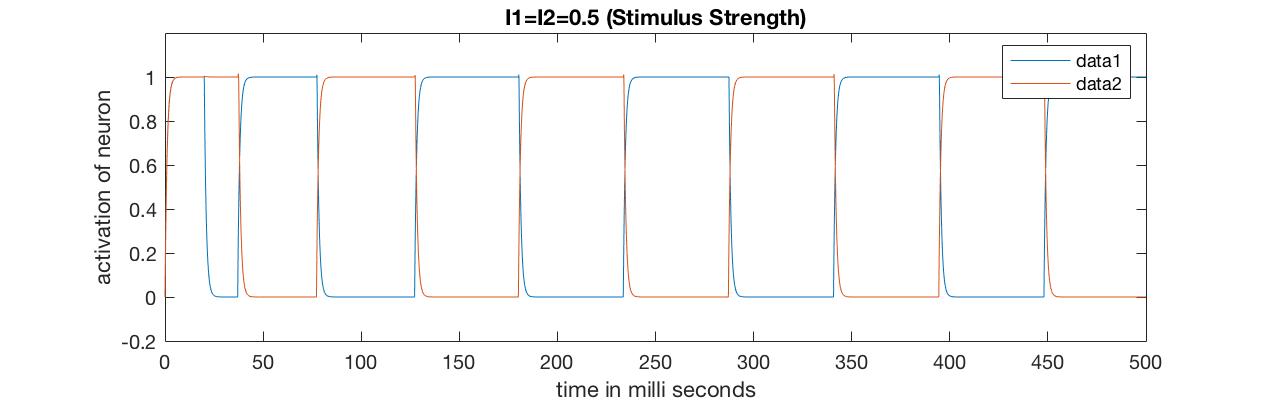}\vspace{1.5mm}
\includegraphics[width=8cm, height=2cm]{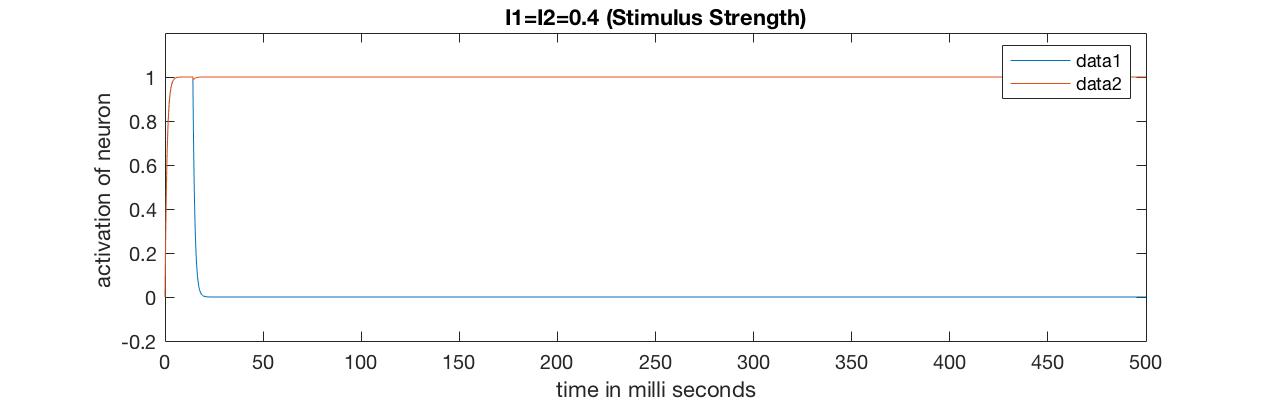}\vspace{1.5mm}
\includegraphics[width=8cm, height=2cm]{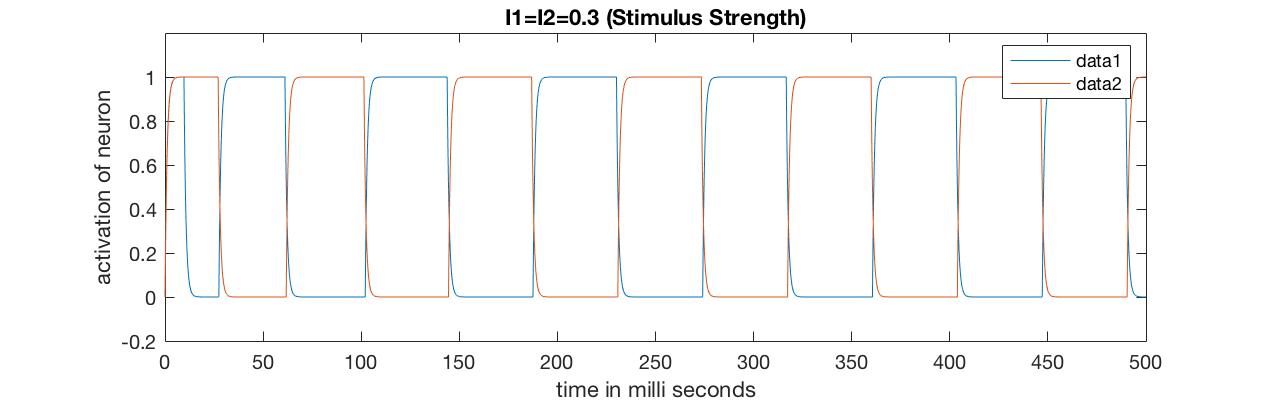}\vspace{1.5mm}
\includegraphics[width=8cm, height=2cm]{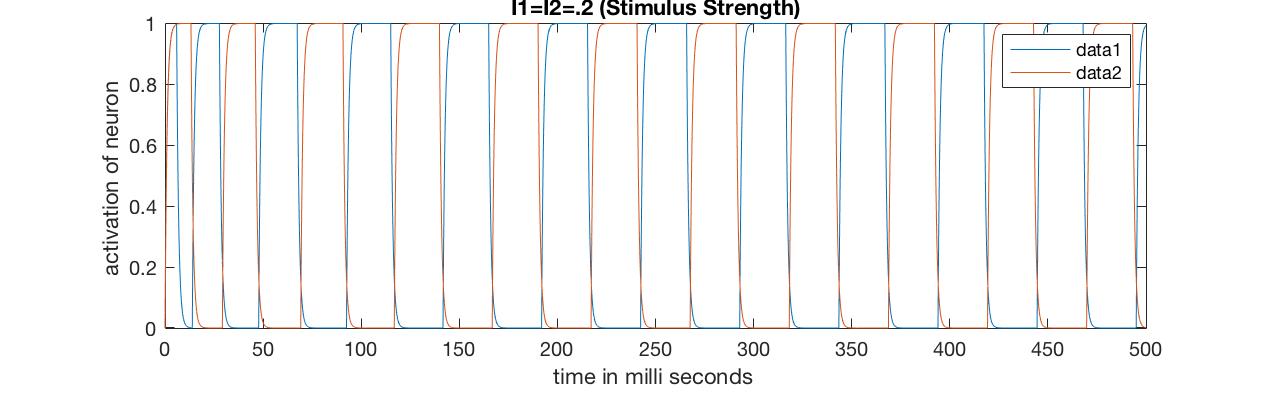}
\label{fig:pic7}
\caption{Laing and Chow's model: Variation of perceptual alternation with stimulus strength. }
\end{figure}

These dynamics do not predict fusion at low stimuli strength or equal dominance at high stimuli strengths. However, there is a winner take all regime (winner determined from the initial conditions) present at intermediate input values. Once again, at high stimuli strengths, as the strength is increased there is a decrease in the mean dominance time while at low stimuli strengths, there is an increase in the mean dominance time. This behaviour is in accordance with Wilson model predictions as well, and once again in contradiction of Levelt's fourth proposition.
\\
To specifically study the effect of adaptation and depression independently on excitatory firing rates of the neuronal populations, two models were derived from Laing and Chow's initial model, namely the adaptation and the depression model \cite{main}.

\subsection{Laing and Chow's Adaptation Model}
\vspace{-2mm}
\begin{equation*}
    \dot{u_1} = -u_1 + f(-\beta u_2 - ga_1 +I_1)
\end{equation*}
\begin{equation*}
    \tau_{a}\dot{a_1} = -a_1 + u_1
\end{equation*}
\begin{equation*}
    \dot{u_2} = -u_2 + f(-\beta u_1 - ga_2 +I_2)
\end{equation*}
\begin{equation*}
    \tau_{a}\dot{a_2} = -a_2 + u_2
\end{equation*}

\subsection{Laing and Chow's Depression model}
\vspace{-2mm}
\begin{equation*}
    \dot{u_1} = -u_1 + f(- \beta u_2g_2 +I_1)
\end{equation*}
\begin{equation*}
    \tau_{d}\dot{g_1} = 1 - g_1 - \gamma u_1g_1
\end{equation*}
\begin{equation*}
    \dot{u_2} = -u_2 + f(- \beta u_1g_1 +I_2)
\end{equation*}
\begin{equation*}
    \tau_{d}\dot{g_2} = 1 - g_2 - \gamma u_2g_2
\end{equation*}

However, in the above two models, $f$ is no longer a Heaviside step function, but is a sigmoid instead. That is $f(x) = 1/{1 + exp[−(x − \theta)/k]}$, where $1/k$ is its slope and $\theta$ is its threshold.\\ 
The adaptation model is simulated for these parameters: $k = 0.1$, $\theta = 0.2$, $g = 0.5$, $\tau_a$ = 100$ms$, and $\beta = 0.9$. Here $g$ is a scaling parameter of the adaptation on neuronal firing rate.  

\begin{figure}[H]
\centering
\includegraphics[width=8cm, height=2cm]{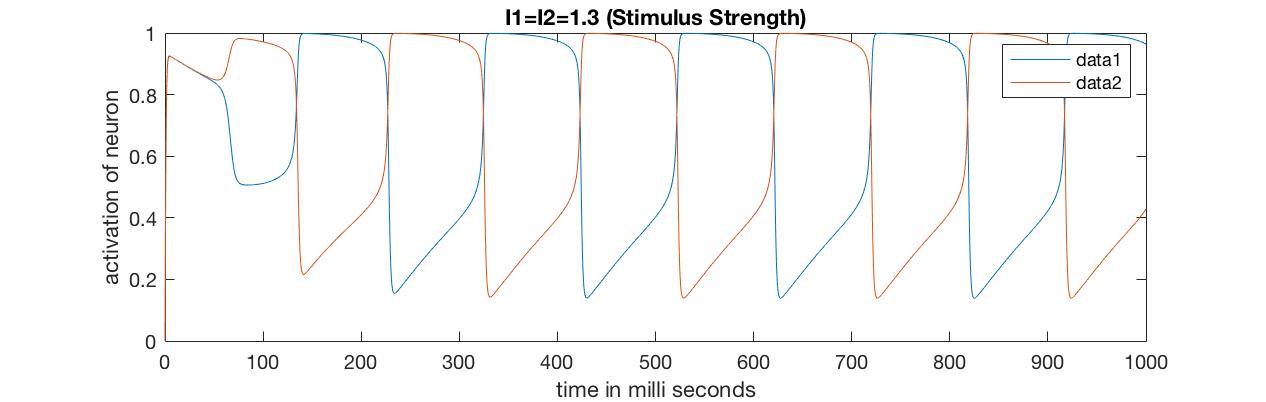}\vspace{2mm}
\includegraphics[width=8cm, height=2cm]{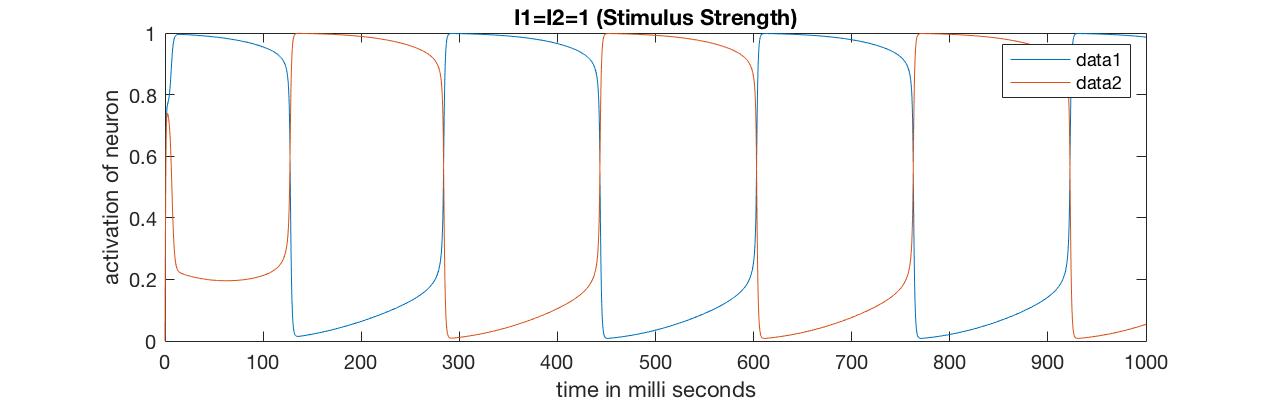}\vspace{2mm}
\includegraphics[width=8cm, height=2cm]{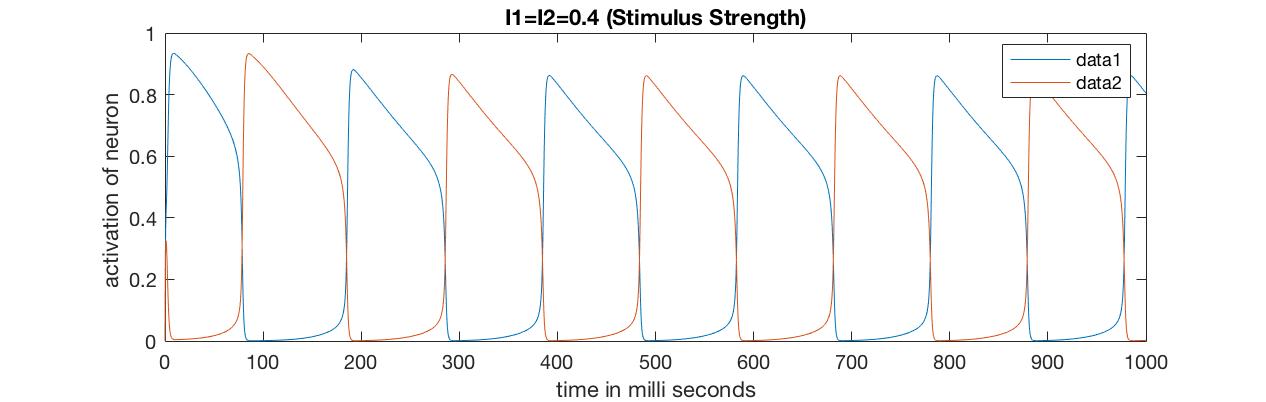}
\end{figure}
\begin{figure}[H]
\centering
\includegraphics[width=8cm, height=2cm]{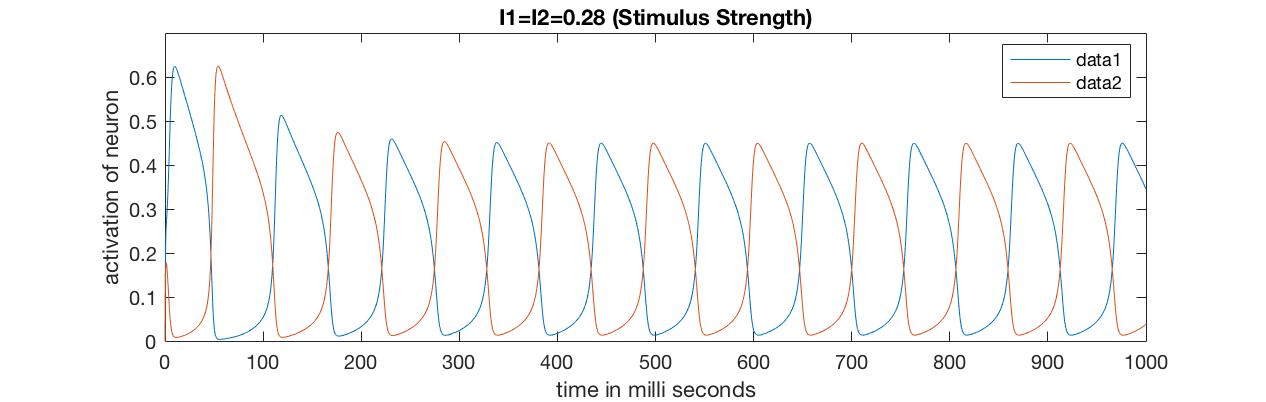}
\caption{Laing and Chow's Adaptaion model: Variation of perceptual alternation with stimulus strength.}
\label{fig:pic7}
\end{figure}

The simulation results of the adaptation model adhere to the previously predicted nonmonotonicity. However, there is an absence of the winner take all regime. This model predicts oscillations for all input stimuli strength. \\
Wilson's model and Laing's models were relatively new in explaining the process of binocular rivalry. A slightly older model is the one proposed by Kalarickal and Marshall. This model shares a few fundamental principles with the above models, however its predictions are quite different. 
\vspace{2mm}
\section{Kalarickal and Marshall's Model}
This model was proposed prior to Wilson's and Laing and Chow's model. This model accounts for adaptation but not synaptic depression and has properties common with Laing's model. $x_i$ is the activation level, $I_i$ is the input stimulus, $W_i^{+}$ is the synaptic weight of the excitatory input pathway to neuron $i$, $W_{ij}^{-}$ is the synaptic weight of the lateral inhibitory pathway from neuron $i$ to neuron $j$, and $y_{ij}$ is the adaptation level of that pathway \cite{kalar}. It is important to notice that this model incorporates noise as well, that is $b(t)$.   
\begin{equation*}
    \dot{x_1} = -x_1 + (1 - x_1)W_1^+I_1 - (c_1 + x_1)W_{21}^-y_{21}max(x_2,0) 
\end{equation*} 
\begin{equation*}
    \dot{y_{21}} = c_2((1-y_{21}) - c_3max(x_2,0)W_{21}^-y_{21}) + b(t)
\end{equation*}
\begin{equation*}
    \dot{x_2} = -x_2 + (1 - x_2)W_2^+I_2 - (c_1 + x_2)W_{12}^-y_{12}max(x_1,0) 
\end{equation*}
\begin{equation*}
    \dot{y_{12}} = c_2((1-y_{12}) - c_3max(x_1,0)W_{12}^-y_{12}) + b(t)
\end{equation*}
The noise variable $b(t)=m$ if $r_{ij}(t) < p$ and $b(t) = -m$ otherwise, where $r_{ij}(t)$ is a uniformly distributed random variable in [0,1), $p$ is a positive constant in [0,1], and $m$ is a small positive constant. For the purpose of simulating the above equations the values of the parameters were taken as $W_1^+ = W_2^+ = 0.25$, $c_1 = 0.01$, $c_2 = 0.008$, $c_3 = 0.083$, $W_{12}^- = W_{21}^- = 250$, $p = 0.5$ and $m = 0.0025$.

\begin{figure}[H]
\centering
\includegraphics[width=8cm, height=2cm]{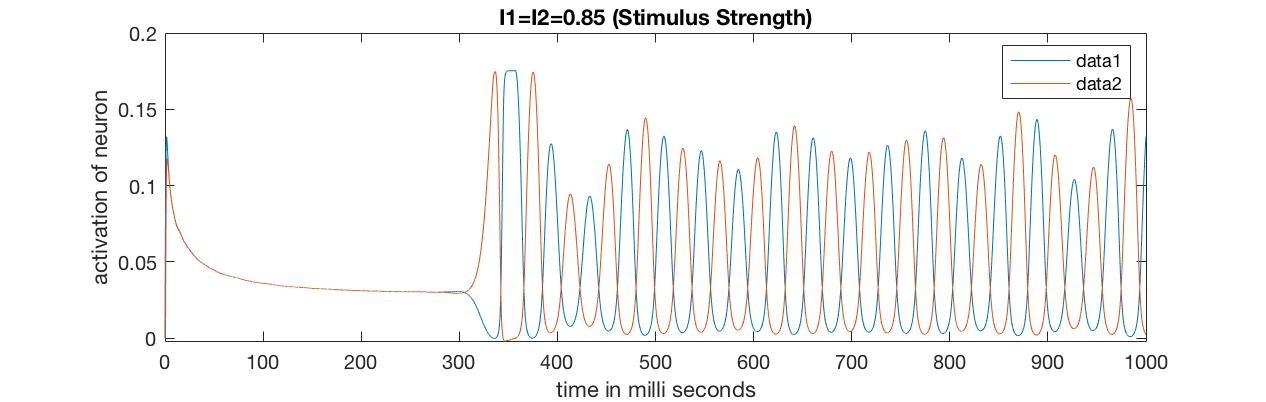}\vspace{2mm}
\includegraphics[width=8cm, height=2cm]{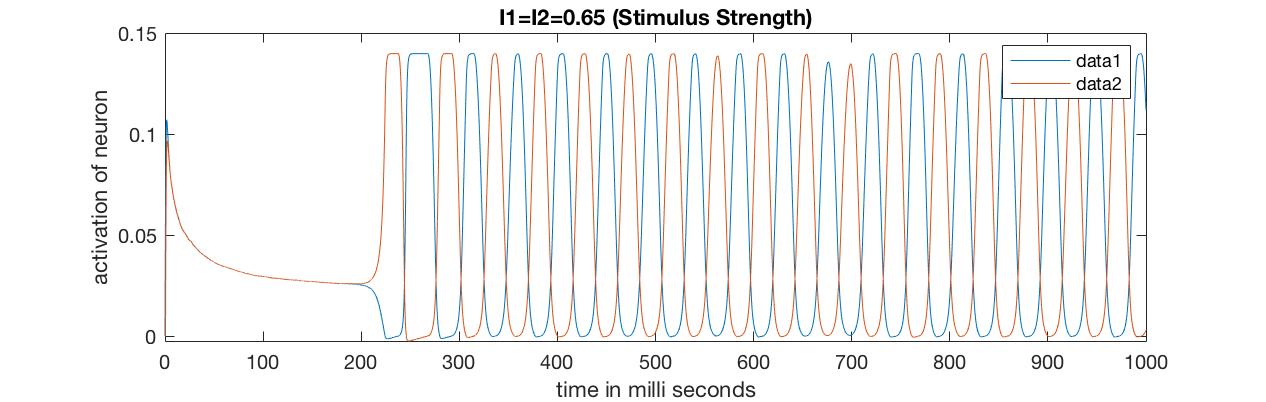}
\end{figure}
\begin{figure}[H]
\centering
\includegraphics[width=8cm, height=2cm]{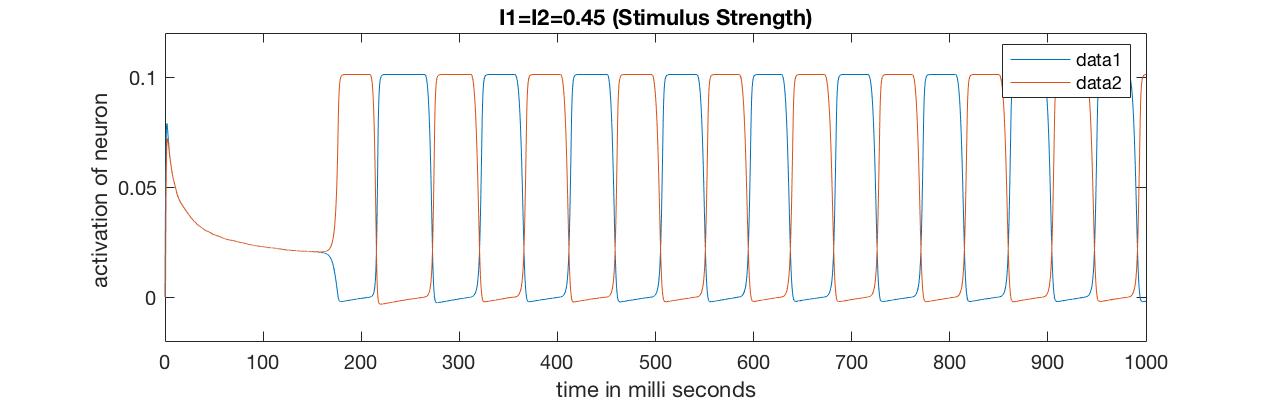}\vspace{2mm}
\includegraphics[width=8cm, height=2cm]{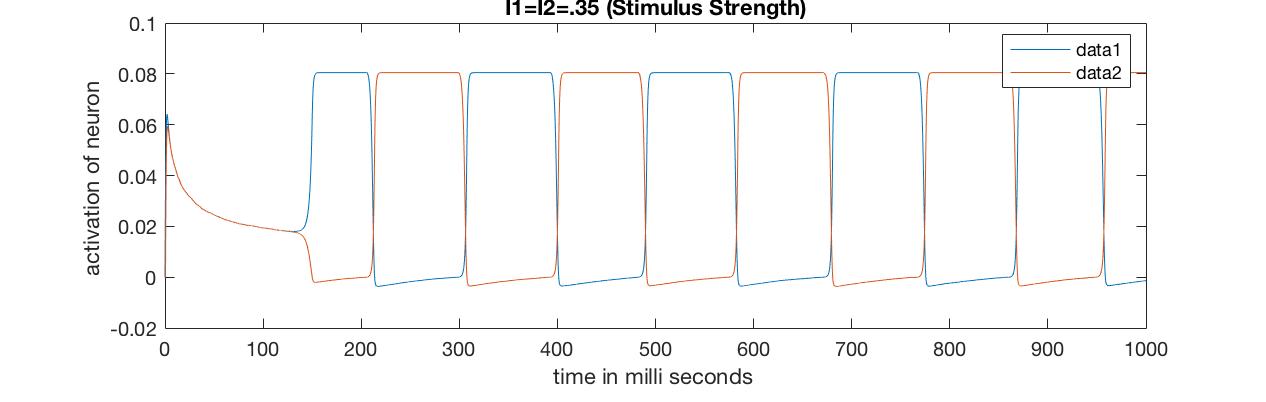}\vspace{2mm}
\includegraphics[width=8cm, height=2cm]{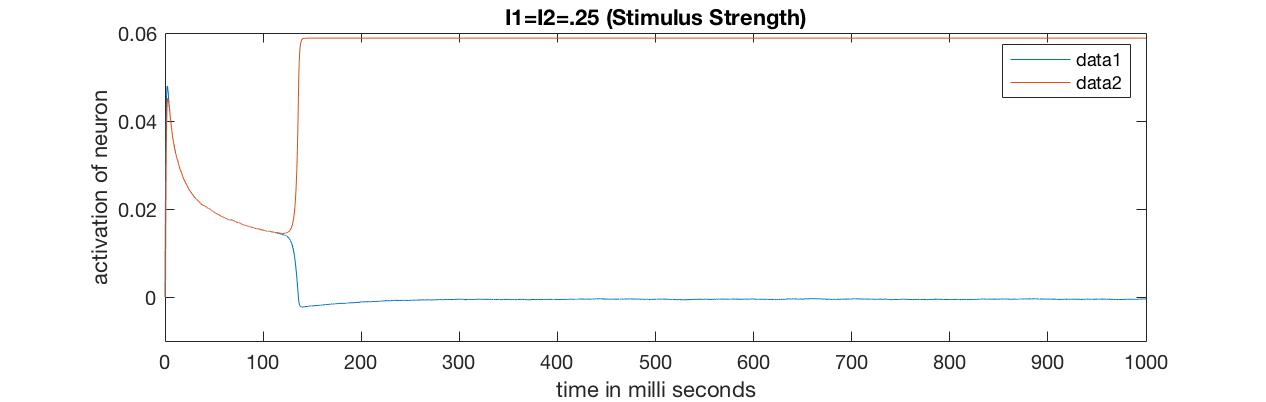}
\caption{Kalarickal and Marshall's model: Variation of perceptual alternation with stimulus strength.}
\label{fig:pic7}
\end{figure}

Kalarickal's model possesses a network architecture similar to that of Laing and Chow's model, that is, a single neuronal population being responsible for visual perception as well as exerting inhibitory effect on the other population. However, its predictions are quite contrary to the above two models. This model predicts no fusion at any stimuli strength. However, at low values there is a dominance of one perception over the other, once again subject to initial conditions. Also, these oscillations are not as uniform as in the previous cases, due to noise being incorporated into the modelling. It also leads to significantly long transition period prior oscillations. However, the trend of the mean dominance time with stimulus strength is robust in the presence of noise as well. It is evident from the simulations, that as stimuli strength increases there is a decrease in the mean dominance time. And this trend is persistent till extremely low stimuli strengths, in agreement with Levelt's fourth proposition.  
\\
Although the models have some differences in their trends, there seems to be an agreement over the existence of oscillations, whose behaviour complies with Levelt's fourth proposition in some range of stimulus strength. The reasons for these oscillations is considered in some detail in the next section, along with discussion on the remaining three of Levelt's propositions.
\vspace{2mm}

\section{DISCUSSION}

When the dominant side is inhibiting the suppressed population from expressing, it is constantly experiencing a slow negative feedback reducing its dominance. And simultaneously, the suppressed population experiences a form of feedback recovery. One possible cause of this flipping in dominance can result from when the negative feedback on the dominant side increases significant enough for its activity to drop below that of the suppressed population, there by letting the latter express itself. This phenomenon is known as "release". In such a phenomenon it is likely to observe a more significant dip in the activity of the active population than a surge in the activity of the suppressed one. Another possible mechanism is known as the "escape" mechanism. This occurs when the feedback recovery of the suppressed population slowly grows enough to result in the overall input becoming positive, thereby resulting in switching of perception. This results in a significant rise in the activity of the suppressed population as when compared to the drop in the activity of the dominant side \cite{main}. The behaviour of these populations just before switching of the dominance is heavily dictated by the gain function incorporated into the modelling, which in the above models has been either Heaviside step function, the sigmoid function or the Naka-Rushton function. The reason for the not allowing negative values for these gain functions, is its resulting repercussion on the population firing rate. Negative values of the gain function results in negative population firing rates, which physically has no significance. 
\\
A salient feature to be observed in the above models is the presence or absence of recurrent excitation. Wilson's, Laing's Adaptation and Kalarickal's models do not incorporate any recurrent excitation where as Laing and Chow's model does. It's evident, that even in its absence, the predictions of Wilson's model are coherent with Laing's to a decent degree. Upon incorporating recurrent excitation into Wilson's model, there was not much change observed in its qualitative behaviour, there by leading to the conclusion that recurrent excitation isn't necessary for the existence of oscillations. In fact, it has been previously studied that it results in system instability and predicts unreported behaviour.\\
Another important consideration is oscillation variation with different parameters. It has been previously found that the qualitative behaviour of these simulations doesn't get altered with differences in the time constants of the slow processes such as adaptation (in case of Wilson's model \cite{main}).  \\
The most important parameter is the one that dictates the strength of cross inhibition. The following simulations were conducted with varying values of the cross inhibition parameter for the same stimulus strength for two models, namely Wilson's and Laing and Chow's.\\

\textit{Laing and Chow's model}
\vspace{-3mm}
\begin{figure}[H]
\centering
\includegraphics[width=8cm, height=2cm]{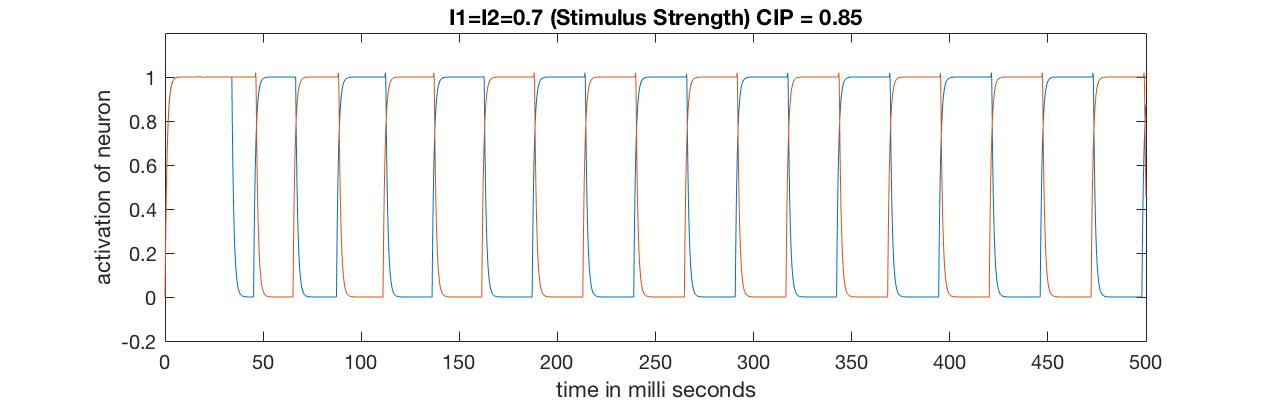}\vspace{2mm}
\includegraphics[width=8cm, height=2cm]{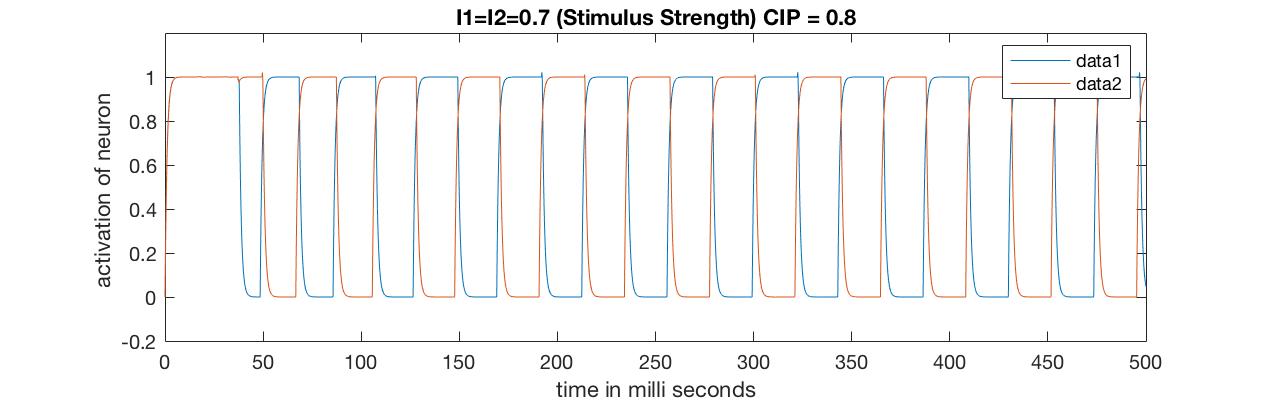}\vspace{2mm}
\includegraphics[width=8cm, height=2cm]{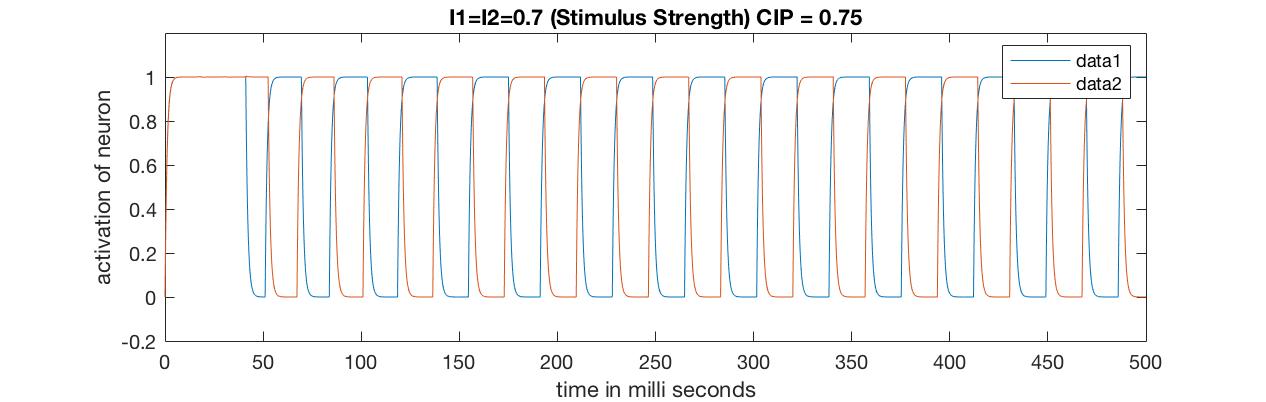}
\end{figure}
\begin{figure}[H]
\centering
\includegraphics[width=8cm, height=2cm]{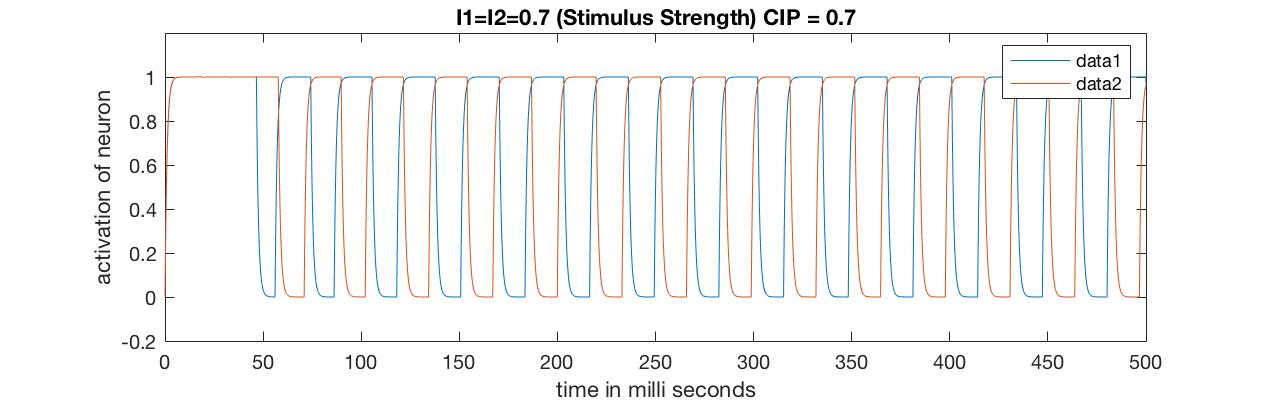}
\caption{Laing and Chow's model: Variation of perceptual alternation with cross inhibition parameter}
\label{fig:pic7}
\end{figure}

\textit{Wilson's Model}
\begin{figure}[h!]
\centering
\includegraphics[width=8cm, height=2cm]{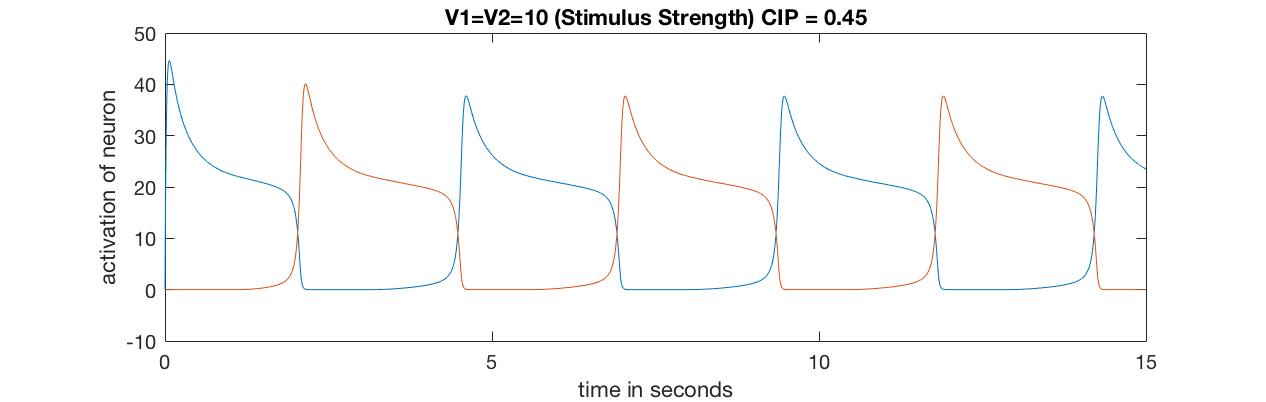}\vspace{2mm}
\includegraphics[width=8cm, height=2cm]{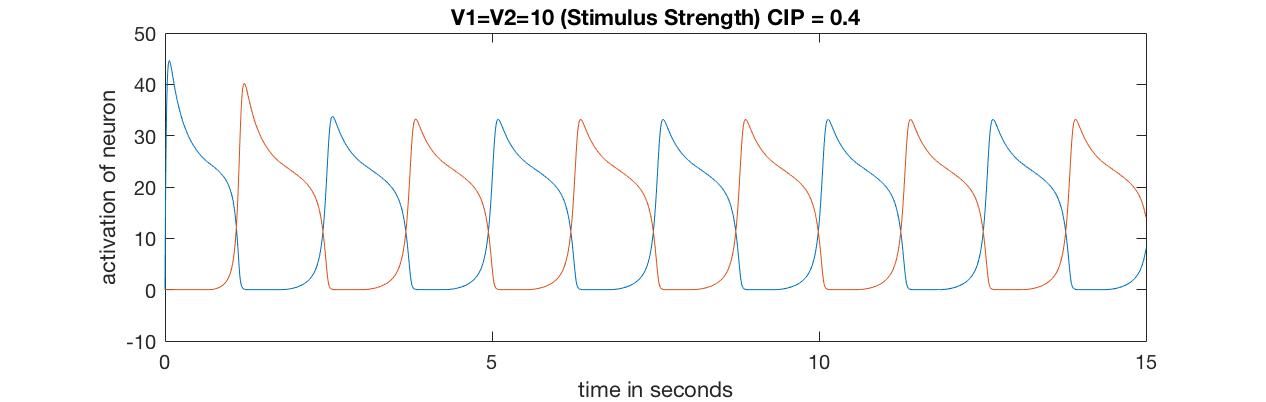}\vspace{2mm}
\includegraphics[width=8cm, height=2cm]{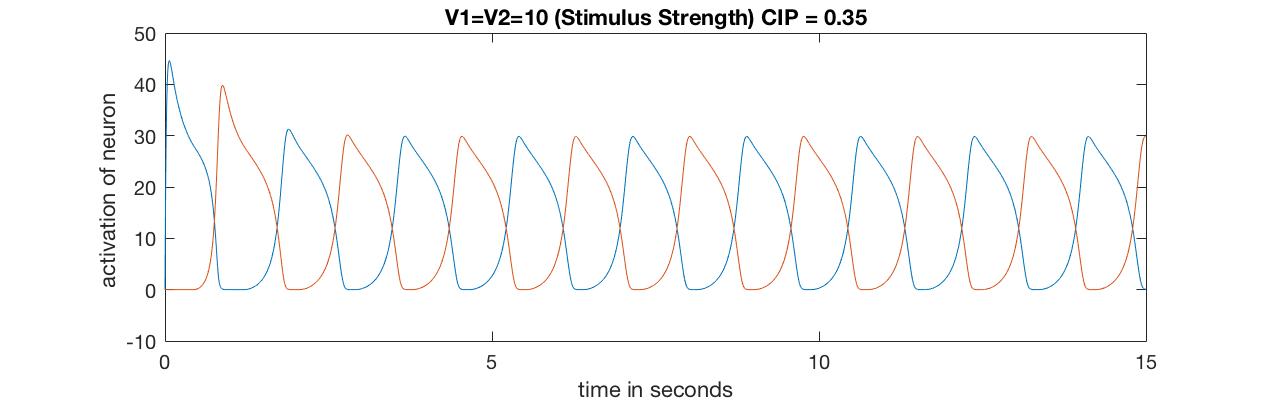}\vspace{2mm}
\includegraphics[width=8cm, height=2cm]{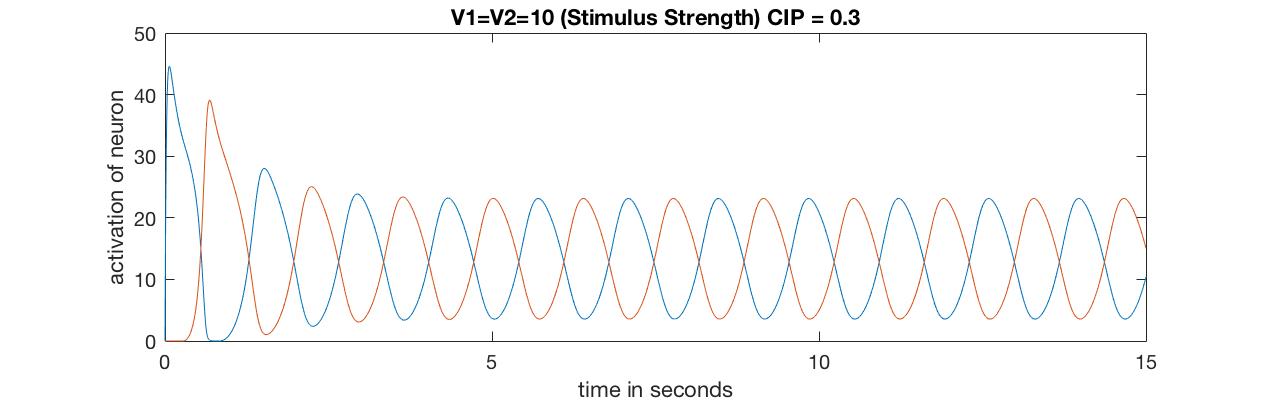}
\caption{Wilson's model: Variation of perceptual alternation with cross inhibition parameter}
\label{fig:pic7}
\end{figure}\\
Both the models present coherent results with respect to the variation of mean dominance time with cross inhibition parameter. As its value is reduced, there is a reduction in the mean dominance time of the perceptual alternations. This phenomenon can be explained from the release and escape perspective presented at the beginning of this section. Lower the value of the cross inhibition parameter, lower the inhibition to be overcome by the suppressed population and lower the negative feedback required by the dominant side to get suppressed. This behaviour is monotonic across multiple stimuli strengths. \\
\\
So far the discussion has been with respect to binocular rivalry in response to stimuli of equal strength. In case of the Wilson model, when simulated for asymmetric stimuli the following is obtained. 
\begin{figure}[H]
\centering
\includegraphics[width=8cm, height=2cm]{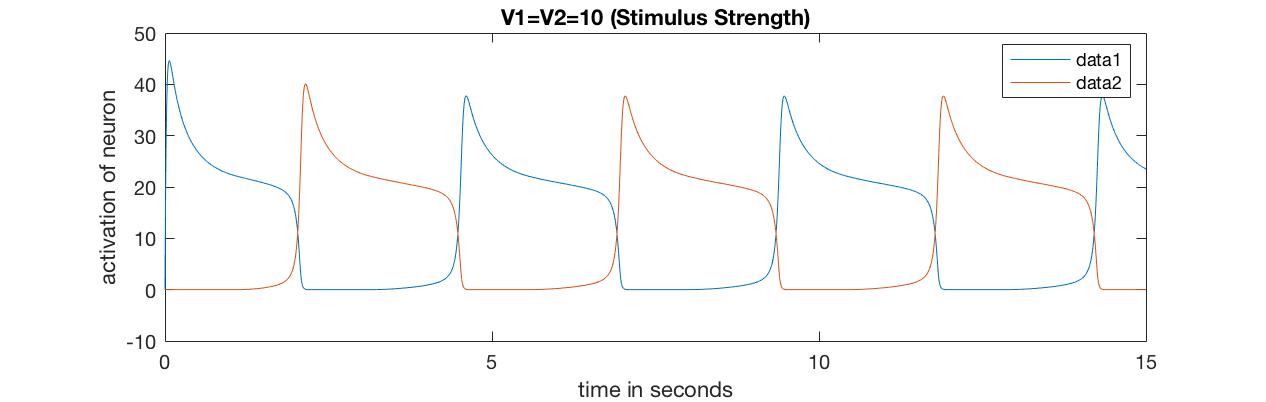}
\end{figure}
\begin{figure}[H]
\centering
\includegraphics[width=8cm, height=2cm]{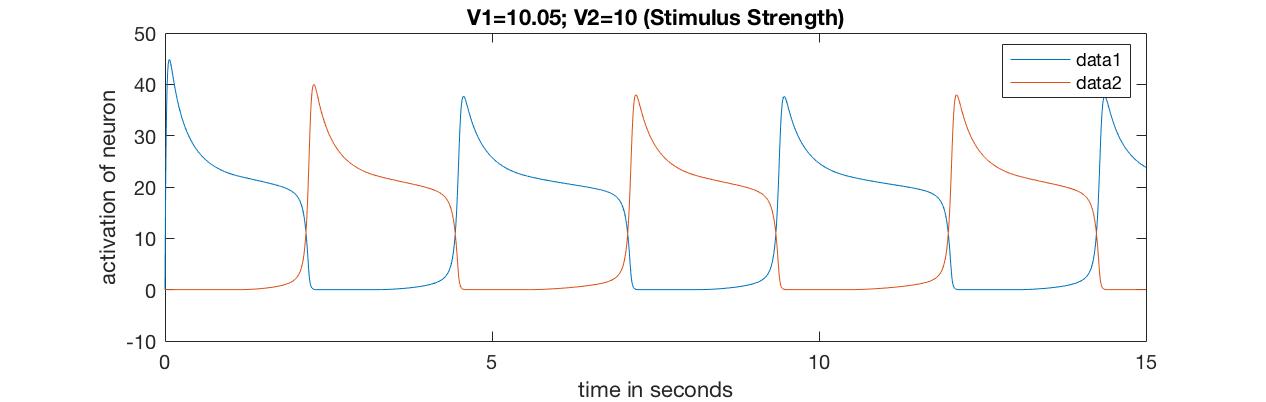}\vspace{2mm}
\includegraphics[width=8cm, height=2cm]{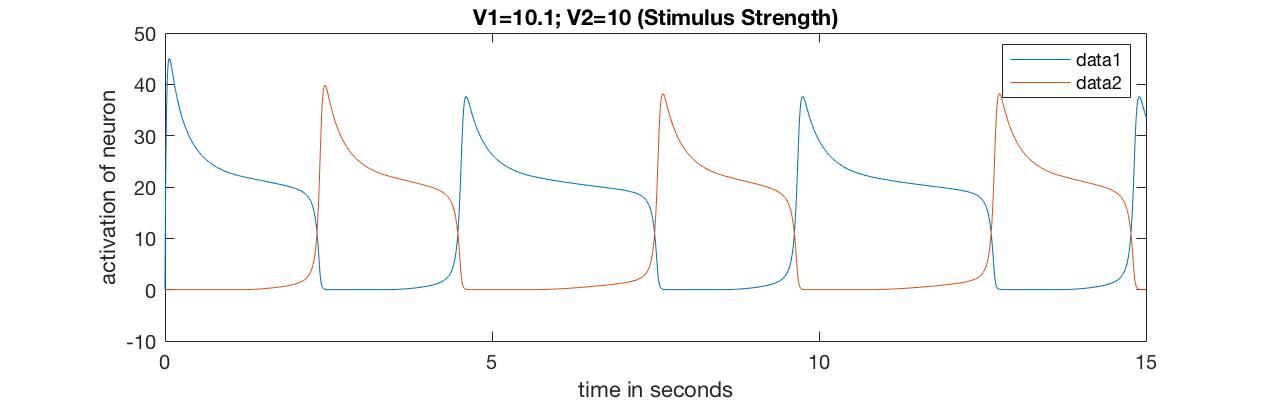}\vspace{2mm}
\includegraphics[width=8cm, height=2cm]{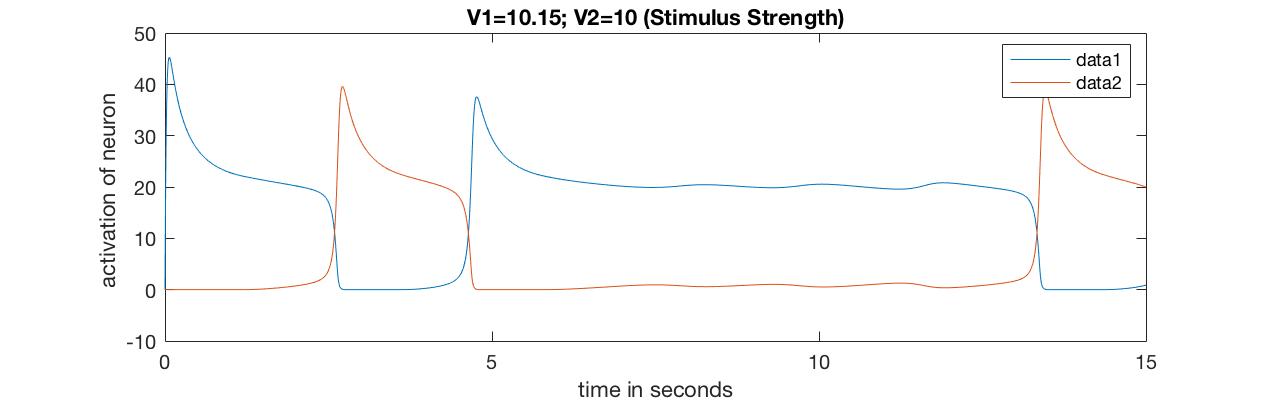}
\caption{Wilson's model: Variation of perceptual alternation with asymmetrical stimuli strength.}
\label{fig:pic7}
\end{figure}
It can be seen that as the strength of one stimulus is increased with respect to the other, the dominance duration of the stronger stimulus increases, while that of the constant stimuli remains the same. This leads to an overall decrease in the rate of the perceptual alternations. This reduction in the alternation rate is highly sensitive to the difference in the stimuli strength. These findings, along with the non monontonic behaviour of mean duration of dominance with varying but equal stimulus strength are in contradiction with the initial propositions of Levelt. They can be altered to accommodate the above as follows.
\\\\
\textit{Modified Levelt's Propositions}
\begin{itemize}
    \item Increasing stimulus strength for one eye will increase the perceptual predominance of that eye's stimulus.
    
    \item Increasing the difference in stimulus strength between the two eyes will primarily act to increase the average perceptual dominance duration of the stronger stimulus.

    \item Increasing the difference in stimulus strength between the two eyes will reduce the perceptual alternation rate.
    
    \item  Increasing stimulus strength in both eyes while keeping it equal between eyes will generally increase the perceptual alternation rate, but this effect may reverse at lower stimulus strengths \cite{levelt}.
\end{itemize}
\vspace{2mm}

\section{Future Work}
The study of rivalry may shed light on neural mechanisms underlying perceptual selection and on the resolution of ambiguous sensory information. Another important question is whether the characteristics reviewed above apply to forms of perceptual bistability outside of vision. An affirmative answer would demonstrate even greater generality of the neural properties that are uncovered by bistable perception paradigms. Similar models can then be used to explain the dynamics of CPG \cite{main} (Central Pattern Generators - biological neural circuits that produce rhythmic outputs in the absence of rhythmic input.) This method of study of rivalry can also be extended to test the presence or absence of auditory or olfactory rivalry. \\
An alteration in the binocular rivalry of an individual can be used in diagnosis of mental disorders such as schizophrenia, bipolar disorder and depression. It has been found that patients suffering with bipolar disorder exhibit lower binocular rivalry rate \cite{disord}. Physiological information on binocular rivalry might aid in diagnosing these diseases more efficiently as well as benefit in the design of relevant treatment and drugs.\\ 
The theoretically important possibility that Proposition IV may not hold true at low stimulus strengths has received only limited experimental verification \cite{levelt}. An advancement in experimental methods and techniques might give better verifiable results. \\  
Recent developments show that a single neural structure is not responsible for the perceptual alternations in each of these phenomena. Further probing can be done into models which account for the spatial distribution of neural structures too. Most of the commonly employed testing equipment involve functional MRIs \cite{fmri} and EEGs. Electrocorticography or intracranial EEGs can aid in better isolation of areas within the brain for binocular rivalry. 

\vspace{2mm}  

%\bibliographystyle{unsrt}%Used BibTeX style is unsrt
%\bibliography{sample}

%\bibliographystyle{unsrt}%Used BibTeX style is unsrt
%\bibliography{root}

\end{document}